\documentclass[conference]{IEEEtran}

 \IEEEoverridecommandlockouts

\usepackage{amssymb,amsmath}
\usepackage{amsthm}
\usepackage{graphicx}
\usepackage{pstricks}
\usepackage{algorithm, algorithmic}

\usepackage{ifthen}
\newboolean{arxiv}
\setboolean{arxiv}{true}

\newtheorem{thm}{Theorem}

\newtheorem{prop}{Proposition}

\theoremstyle{definition}
\newtheorem{defn}{Definition}

\newcommand{\abs}[1]{\left \lvert #1 \right \rvert}
\newcommand{\absS}[1]{ \lvert #1 \rvert}
\newcommand{\norm}[1]{\left \lVert #1 \right\rVert}

\newcommand{\st}{\;:\;}                         

\title{Parallel coordinate descent for the Adaboost problem}

\author{\IEEEauthorblockN{Olivier Fercoq\thanks{The work of the author was supported by the EPSRC grant EP/I017127/1 (Mathematics for Vast Digital Resources).}
} \IEEEauthorblockA{School of Mathematics\\ University of Edinburgh\\ United Kingdom} }

\begin{document}

\maketitle

\begin{abstract}
We design a randomised parallel version of Adaboost based on previous studies on parallel coordinate descent. The algorithm uses the fact that the logarithm of the exponential loss is a function with coordinate-wise Lipschitz continuous gradient, in order to define the step lengths. We provide the proof of convergence for this randomised Adaboost algorithm and a theoretical parallelisation speedup factor. We finally provide numerical examples on learning problems of various sizes that show that the algorithm is competitive with concurrent approaches, especially for large scale problems.
\end{abstract}

\section{Introduction}

The Adaboost algorithm, introduced by Freund and Shapire~\cite{ freund1997decision}, is 
a widely used classification algorithm. Its goal is to combine many weak hypotheses with high error rate to generate a single strong hypothesis with very low error.
The algorithm is equivalent to the minimisation of the exponential loss by the greedy coordinate descent method~\cite{schapire1999improved}. At each iteration, it selects the classifier with the largest error
and one updates its weight in order to decrease at most this error. The weights of the other classifiers are left unchanged.

Adaboost has found a large number of applications and to name a few, we may cite
face detection and recognition~\cite{viola2001rapid}
 cancer detection by interpretation of radiographies~\cite{zinovev2011lungadaboost}, 
gene-gene interaction detection~\cite{assareh2012geneadaboost}\ldots

The original algorithm is intrinsically sequential, but several parallel versions have been developped.

Collins, Shapire and Singer~\cite{collins2002logistic} give a version of the algorithm where all the coordinates are updated in parallel. They prove the convergence of this fully parallel coordinate descent under the assumption that the 
1-norm of each row of the feature matrix has a norm smaller than 1. One may relax this assumption by requiring that every element of the matrix has absolute value smaller than 1 and by dividing the step length by the maximum number of nonzero elements in a row that we will denote $\omega$.
 
In the context of support vector machines, Mukherjee et al.~\cite{mukherjee2013parallel} interpreted
the  fully parallel coordinate descent method as a gradient method and designed an accelerated 
algorithm using accelerated gradient. The same approach is possible for Adaboost and we give numerical experiments
in Section~\ref{sec:numexp}.

Another approach for parallelisation  is proposed in~\cite{zeng2011parallization}: the author 
keep the Adaboost algorithm unchanged but parallelise the inner operations.

Finally, \cite{merler2007parallelizing} proposed to solve a different problem, that will give a similiar result 
to Adaboost but will be solved by a parallel algorithm. However, they need to initialise the algorithm with
iterations of the sequential Adaboost and they only give empirical evidence that the number of iterations required is small. 
 
Palit and Reddy~\cite{palit2012scalable} first partition the coordinates so that each processor get a subset of the data.
Each processor solves the Adaboost problem on its part of the data and the results are then merged. The algorithm
involves very few communication between processors but the authors only provided
a proof of convergence in the case of two processors. 

In this paper, we propose a new parallel version of Adaboost based on recent work on parallel coordinate descent.
In~\cite{RT:PCDM}, Richt\'arik and Tak\'a\v{c} introduced a general parallel coordinate descent method for
the minimisation of composite functions of the form
$F(x)=f(x)+\psi(x)$,
where $f$ is convex, partially separable of degree $\omega$ and has a coordinate-wise Lipschitz gradient, and $\psi$ is
a  convex, nonsmooth and separable function, e.g.\ the $l_1$ norm.
They provided convergence results for this algorithm together with a theoretical parallelisation speedup factor.
They obtained the best speedups for randomised coordinate descent methods, which means that 
the coordinates are chosen according to a random sampling of $\{ 1, \ldots, n\}$.
They showed~\cite{takac2013mini} that this algorithm is very well suited to the resolution of Support Vector Machine problems.

The exponential loss does not fit in this framework because it does not have a Lipschitz gradient.
However, Fercoq and Richt\'arik~\cite{FR:spcdm} showed that the parallel coordinate descent method
can also be applied in the context of nonsmooth functions with max-structure to so called Nesterov separable functions.

We show in Theorem 1 that the logarithm of the exponential loss is Nesterov separable and this allows us
to define the parallel coordinate descent method for the Adaboost problem (Algorithm \ref{alg:paralleladaboost}). 
Then, we prove the convergence of the algorithm (Theorem~\ref{thm:convergence}) and give its iteration complexity, basing on the iteration complexity of the classical Adaboost \cite{mukherjee2011rate, telgarsky2012primal}.
Finally, we provide numerical examples on learning problems of various sizes.

\section{The Adaboost problem}

Let $M \in \mathbb{R}^{m \times n}$ be a matrix of features and $y \in \mathbb{R}^m$ be a vector of labels.
We denote by $A\in \mathbb{R}^{m \times n}$ the matrix such that
\[
A_{j,i}=y_j M_{j,i} \enspace.
\]
In this paper, we may accept $A_{j,i} \not \in [-1, 1]$.
We will write the coordinates as indices and the sequences as superscripts. Hence $\lambda^t_i$ is the 
$i^{th}$ coordinate of the $t^{th}$ element of the vector valued sequence $(\lambda^t)_{t\geq 0}$.

The Adaboost problem is the minimisation of the exponential loss~\cite{friedman2000additive}:
\begin{equation} \label{eq:adaprob}
\min_{\lambda \in \mathbb{R}^n} \frac{1}{m} \sum_{j=1}^m \exp( (A \lambda)_j) .
\end{equation}

Let $f : \mathbb{R}^m \to \mathbb{R}$ be the following
empirical risk function
\[ 
f(x)=\frac{1}{m} \sum_{j=1}^m \exp(x_j) .
\]
We denote the optimal value of the Adaboost problem \eqref{eq:adaprob} by
\[
\bar{f}_A=\inf_{\lambda \in \mathbb{R}^m} f(A \lambda).
\]

It will be convenient to consider the following equivalent objective function with Lipschitz gradient
\[
F(\lambda)=\log (f(A \lambda)) ,
\]
and its  associated $C^{1,1}$ Adaboost problem
\begin{equation} \label{eq:adaprobC1}
\min_{\lambda \in \mathbb{R}^n} F(\lambda) .
\end{equation}
As the logarithm is monotone, problems \eqref{eq:adaprob} and \eqref{eq:adaprobC1} are equivalent.
Moreover both are convex optimisation problems. 
This version of the Adaboost problem has a nice dual problem involving the entropy function~\cite{shen2010dual} 
and  the Lipschitz continuity of the gradient of $F$ 
will be useful to define the Parallel Coordinate Descent Method.

\section{Parallel coordinate descent}

\subsection{General case}

In this section, we present the Parallel Coordinate Descent Method 
introduced by Richt\'arik and Tak\'a\v{c} in~\cite{RT:PCDM}.

For $w \in \mathbb{R}_{++}^n$, we denote by $\norm{\cdot}_w$ the norm such 
that $\norm{x}_w=\big( \sum_{i=1}^n w_i (x_i)^2 \big)^{1/2}$.

At each iteration of the parallel coordinate descent method, one needs to select
which coordinates will be updated. 
One may choose the coordinates to update in
a given deterministic way but it is convenient to
randomise this choice of variables.
Several samplings, i.e., laws for randomly choosing subsets of variables of  $\{1, \ldots, n\}$, are considered in~\cite{RT:PCDM}.

We will focus in this paper on the $\tau$-nice sampling $\hat{S}$. 
It corresponds to the case
where we have $\tau$ processors updating $\tau$ coordinates
in parallel and each subset of $\{1, \ldots, n\}$ with $\tau$ coordinates has the same probability to be selected:
\[
\mathbf{P}(\hat{S}=S) = \begin{cases} \frac{1}{\binom{n}{\tau}} , &\text{ if } \abs{S}=\tau \\ 0 , & \text{otherwise.}\end{cases} 
\]

A good approximation of the $\tau$-nice sampling for $\tau \ll n$ is the $\tau$-independent
sampling where each processor selects the coordinate it will update following a  uniform law,
independently of the others. 

The choice of the sampling has consequences on the complexity estimates.
More precisely,
the parallel coordinate descent method relies on the concept of
Expected Separable Overapproximation (ESO) to compute the
updates and the ESO depends on the sampling. We denote here by $h_{[S]}$ the vector of $\mathbb{R}^n$ 
such that $(h_{[S]})_i=h_i$ if $i \in S$ and $(h_{[S]})_i=0$ otherwise.
\begin{defn}[\cite{RT:PCDM}]
 Let $\beta>0$, $w \in \mathbb{R}^n_{++}$ and $\hat{S}$ be a sampling.
We say that $f: \mathbb{R}^n \to \mathbb{R}$ admits a {\em $(\beta,w)$-Expected Separable Overapproximation} with respect
to $\hat{S}$ if for all $x,h \in \mathbb{R}^n$,
\begin{equation} 
\label{eq:eso}
 \mathbf{E}[F(x+h_{[\hat{S}]})] \leq F(x) + \frac{\mathbf{E}[\absS{\hat{S}}]}{n} \Big( \langle \nabla F(x), h \rangle + \frac{\beta}{2} \norm{h}_w^2 \Big) .
\end{equation}
We denote $(F, \hat{S})\sim$ESO$(\beta,w)$ for simplicity.
\end{defn}

As the overapproximation is separable, one can find a minimiser
with respect to $h$ by $n$ independent optimisation problems
that will return $h_i$ for $i \in \{ 1, \ldots ,n \}$.
In fact, we do not even need to compute the coordinates of $h$ that are not needed afterwards,
i.e.\ we only compute $h_i$ for $i \in \hat{S}$.

\begin{algorithm}
  \begin{algorithmic}
\STATE Compute $\beta$ and $w$ such that $(F, \hat{S}) \sim \mathrm{ESO}(\beta,w)$.
\FOR{$t \geq 0$}
 \STATE Randomly generate $S^t$ following sampling $\hat{S}$.
 \STATE Compute $h_i$, $i \in S^t$ where $h$ minimises the overapproximation~\eqref{eq:eso}.
\STATE $x^{t+1} \leftarrow x^t + h_{[S^t]}$
\IF{$F(x^{t+1})>F(x^t)$}
 \STATE $x^{t+1} \leftarrow x^t$
\ENDIF
\ENDFOR
  \end{algorithmic}
\caption{Parallel Coordinate Descent Method \cite{RT:PCDM}}
\label{alg:pcdm2}
\end{algorithm}

The convergence properties of the Parallel Coordinate Descent Method (Algorithm \ref{alg:pcdm2})
have been studied  for quite general classes of functions, namely
partially separable functions~\cite{RT:PCDM} and Nesterov separable functions~\cite{FR:spcdm}.
The addition of a separable regulariser like the $l_1$-norm or box constraints was also considered.
However, in all cases, the analysis assumes that there exists a minimiser, which is not true in general
for the Adaboost problem \eqref{eq:adaprob}.

\subsection{Adaboost problem}

In the following, we specialise the Parallel Coordinate Descent Method to the $C^{1,1}$ Adaboost problem~\eqref{eq:adaprobC1}.
We begin by giving an ESO for the logarithm of the objective function.

\begin{thm}
\label{thm:spcdm}
Let $\omega$ be the maximum number of element in a row of matrix $A$, that is
\[
\omega= \max_{1 \leq j \leq m} \abs{ \{ i \in \{1, \ldots, n\} \st A_{i,j} \not = 0\} } .
\]
Let us denote
\[
 p_l=\frac{\binom{\omega}{l} \binom{n-\omega}{\tau-l}}{\binom{n}{\tau}} , \; c_l=\max\Big(\frac{l}{\omega}, \frac{\tau -l}{n-\omega}\Big)
 , \; 0 \leq l \leq \min(\omega,\tau)
\]
($c_l=l/\omega$ if $\omega=n$) and 

\vspace{-3ex}

\begin{equation*}
\beta=\sum_{k=1}^{\min(\omega, \tau)} \!\! \min \Big (1, \frac{mn}{\tau} \sum_{l=k}^{\min(\omega,\tau)}c_l p_l \Big) \enspace .
\end{equation*}
The function $F$ has a coordinate-wise Lipschitz gradient with constants $(L_i)_{1 \leq i \leq n}$ such that
\[
L_i = \max_{1 \leq j \leq m} A_{j,i}^2 \enspace,
\]
and if one chooses a $\tau$-nice sampling $\hat{S}$, then 
\[
(F,\hat{S})\sim~\mathrm{ESO}(\beta,L) \enspace.
\]
\end{thm}
\begin{IEEEproof}
By \cite{Nesterov05:smooth}, Section 4.4, we know that  $F$  can be written as
\begin{align*}
F(\lambda)&=\log(\frac{1}{m}\sum_{j=1}^m \exp( (A\lambda)_j) \\
&= \max_{u \in \Sigma_m} \langle  A \lambda , u \rangle - d_2(u)-\log(m),
\end{align*}
where $\Sigma_m=\{u \in \mathbb{R}^m \st u\geq 0 , \; \sum_{j=1}^m u_j =1 \}$ is the simplex of $\mathbb{R}^m$ and 
$
d_2(u) = \sum_{j=1}^m u_j \log(u_j) + \log(m)
$
is 1-strongly convex on $\Sigma_m$ for the 1-norm.

This shows that $F$ is Nesterov separable of degree $\omega$ in the sense of~\cite{FR:spcdm}, and so the Lipschitz constants are given by Theorem 2 in~\cite{FR:spcdm}. Moreover, by Theorem 6 in \cite{FR:spcdm},  if one chooses a $\tau$-nice sampling $\hat{S}$, $(F,\hat{S})\sim$ ESO$(\beta,L)$.
\end{IEEEproof}

We can now state the Parallel Coordinate Descent Method for the $C^{1,1}$ Adaboost problem \eqref{eq:adaprobC1},
since the minimiser of the ESO for $F$ and a $\tau$-nice sampling $\hat{S}$ at $\lambda$ is
given by $\delta \in \mathbb{R}^n$ such that for all $i$,  $\delta_i =\frac{1}{\beta L_i} \nabla_i F(\lambda)$.
\begin{algorithm}
  \begin{algorithmic}
\STATE Compute $\beta$ and $(L_i)_{1\leq i \leq n}$ as in Theorem \ref{thm:spcdm}.
\FOR{$t \geq 0$}
 \STATE Randomly generate $S^t$ following sampling $\hat{S}$.
\renewcommand{\algorithmicdo}{\textbf{do in parallel}}
 \FOR{$i \in S^t$}
    \STATE $\delta_i \leftarrow \frac{1}{\beta L_i} \nabla_i F(\lambda^t)$
     \STATE $\lambda_i^{t+1} \leftarrow \lambda^t_i + \delta_i$
\ENDFOR
\renewcommand{\algorithmicdo}{\textbf{do}}
\IF{$F(\lambda^{k+1})>F(\lambda^k)$}
 \STATE $\lambda^{k+1} \leftarrow \lambda^k$
\ENDIF
\ENDFOR
  \end{algorithmic}
\caption{Parallel Adaboost}
\label{alg:paralleladaboost}
\end{algorithm}

\subsection{Computational issues}

Computation of $(L_i)_{1\leq i \leq n}$ is easy and can be done with one single read of the data. 

For $\beta$, we shall first compute $p_l$ for $l \in \{1, \ldots, \min(\omega, \tau)\}$. Note that
\begin{align*}
p_l&=\frac{\binom{\omega}{l} \binom{n-\omega}{\tau-l}}{\binom{n}{\tau}}  
 = \frac{(n-\omega) \ldots (n-\omega-\tau+l+1)}{n \ldots (n-\tau+l+1)}\\
& \qquad \qquad \times\frac{\omega \ldots (\omega-l+1)}{(n-\tau+l)\ldots (\tau+1)} 
 \frac{\tau \ldots (\tau-l+1)}{l \dots 2.1}
\end{align*}
There are $(\tau-l)+l+l=\tau+l$ divisions of integers and the multiplication of these terms. Paired as in the
last expression, none of the terms to multiply is bigger than $(\tau-l+1)$ and with a reshuffling of the terms 
before the multiplication, one can easily get a numerically stable way of computing $p_k$. Then, we just need to perform
simple sums and comparisons with 1.

The gradient of $F$ is given by
\[
\nabla F(\lambda)= p(\lambda)^T A,
\]
where for all $j \in \{ 1, \ldots, m\}$, 
\[
p_j(\lambda)=\frac{\exp((A\lambda)_j)}{\sum_{k=1}^m \exp((A \lambda)_k)}
\]

To compute the gradient, one stores residuals $r_j=(A \lambda)_j$ and updates them at each iteration,
as well as the function values $f(A \lambda)=\sum_{k=1}^m \exp((A \lambda)_k)$. If we start with $\lambda^0=0$,
there is no big number in $f(A \lambda)$. The value of the function can be updated in parallel by a reduction clause
and used both for the computation of the gradient and the test in Algorithm~\ref{alg:paralleladaboost}.

\section{Convergence of parallel coordinate descent for the Adaboost problem}

The proof of convergence follows the lines of~\cite{telgarsky2012primal} with additional technicalities due to the 
randomisation of the samplings and the introduction of the logarithm. For conciseness of this paper,
we give the proof and the precise definition of the parameters in
\ifthenelse{\boolean{arxiv}}{the appendix.}{an extended version~\cite{Fercoq-paralleladaboostextended}.} 

 Theorem~\ref{thm:convergence}
 gives a bound on the number of iterations needed for the Parallel Coordinate Descent Method (Algorithm~\ref{alg:paralleladaboost}) to return with high probability, an $\epsilon$-solution to the Adaboost problem~\eqref{eq:adaprob}. 

\begin{thm}
\label{thm:convergence}
Suppose $1 \leq \abs{H(A)}\leq m-1$. Partition the rows of $A$ into $A_0 \in \mathbb{R}^{m_0 \times n}$ and $A_+ \in \mathbb{R}^{m_+ \times n}$, and suppose the axes of $\mathbb{R}^m$ are ordered so that $A=\begin{bmatrix} A_0 \\ A_+\end{bmatrix}$. Set $\mathcal{C}_+$ to be the tightest axis-aligned rectangle such that $\{ x \in \mathbb{R}^{m_+} \st (f+I_{Im(A_+)})(x) \leq f(A \lambda^0)\} \subseteq \mathcal{C}_+$, and $\tilde{w} = \sup_{t \geq 0} \frac{1}{f(A_+ \lambda^t)} \norm{\nabla f(A_+ \lambda^t) - \mathrm{P}^1_{\nabla f(\mathcal{C}_+) \cap \mathrm{Ker}(A_+^T)}(\nabla f (A_+ \lambda^t))}_1$.
Then $\mathcal{C}_+$ is compact, $w < +\infty$, $\log \circ f$ has modulus of strong convexity $\tilde{c}>0$ over $\mathcal{C}_+$ and 
$\tilde{\gamma}(A, \mathbb{R}^{m_0} \times \nabla f(\mathcal{C}_+)) >0$. 

Using these terms,
choose an initial point $\lambda^0$, an accuracy $\epsilon$, $0 < \epsilon < 2 \bar{f}_A$, a confidence level $\rho>0$ and iteration counter
\[
T \geq  
 \frac{4 \beta n}{\tau} \frac{(1+2\tilde{w}/\tilde{c})^2}{\tilde{\gamma}(A,\mathbb{R}_+^{m_0} \times \nabla f(\mathcal{C}_+)^2} \frac{f(A \lambda^0)^2}{\bar{f}_A}
\frac{1}{\epsilon} (1 + \log \frac{1}{\rho})+2  \enspace.
\]
Then the $T^{th}$ iterate $\lambda^T$ of the Parallel Coordinate Descent Method (Algorithm~\ref{alg:paralleladaboost}) applied to $F(\lambda)=\log(f(A\lambda))$
with a $\tau$-nice sampling is an $\epsilon$-solution to the original Adaboost problem \eqref{eq:adaprob} with probability at least $1-\rho$:
  \[
\mathbf{P}(f(A \lambda^T)-\bar{f}_A \leq \epsilon)\geq 1-\rho \enspace.
\]
\end{thm}

The iteration complexity is in $O(1/\epsilon)$, like for the classical Adaboost algorithm~\cite{mukherjee2011rate, telgarsky2012primal}. The theorem also gives the theoretical parallelisation speedup factor of the method, which is
equal to~$\frac{\tau}{\beta}$, where $\beta$ is given in Theorem~\ref{thm:spcdm} and is always smaller than~$\min(\omega,\tau)$. This means that, when one neglects communication costs, the algorithm is $\frac{\tau}{\beta}$ faster with $\tau$ processors than with one processor.
The value of $\beta$ can be significantly smaller than $\min(\omega,\tau)$ when using a $\tau$-nice sampling. For instance 
for the experiment on the URL reputation dataset, $\beta \approx 3.2$ when $\tau=16$.

We now give the convergence results in the case of weak learnability and attainability.

\begin{prop}
\label{prop:weaklearn}
If $\abs{H(A)}=0$, choosing
\[
T \geq \frac{\beta}{\tau }\frac{2n}{\tilde{\gamma}(A, \mathbb{R}_+^m)^2}\log\Big( \frac{f(A\lambda^0)}{\epsilon \rho} \Big)
\]
grants
\[
\mathbf{P}(f(A\lambda^t) \leq \epsilon) \geq 1-\rho.
\]
\end{prop}

\begin{prop}
\label{prop:attain}
If $\abs{H(A)}=m$, choosing $0< \epsilon < 2 \bar{f}_A$ and
\[
T \geq \frac{\beta}{\tau} \frac{4n}{ \tilde{c} \tilde{\gamma}(A, \nabla f(\mathcal{C}))^2}\log\Big( 2\frac{f(A\lambda^0)-\bar{f}_A}{\epsilon \rho} \Big)
\]
grants
\[
\mathbf{P}(f(A\lambda^t) \leq \epsilon) \geq 1-\rho.
\]
\end{prop}

More iterations are needed than with the greedy update but here we do not need to find the coordinate of the gradient
with the biggest absolute value, which saves computational effort. However, in both cases, the parameters $\gamma, w, c$ and $\tilde{\gamma}, \tilde{w}, \tilde{c}, \bar{f}_A$ are not easily computable. Hence, the biggest interest of these convergence results is to show that the convergence holds in $\frac{1}{\epsilon}$ (or $\log \frac{1}{\epsilon}$) for any initial point.

\section{Numerical experiments}
\label{sec:numexp}

In this section, we compare the Parallel Coordinate Descent Method with three other algorithms available for the resolution
of the Adaboost problem. We will not consider algorithms that solve a different problem like the ones presented in~\cite{merler2007parallelizing, palit2012scalable}.

We run our experiments on two freely available datasets: w8a~\cite{platt199912} and URL reputation~\cite{ma2009identifying}.
The w8a dataset is of medium scale: it has $m=49749$ examples, $n=300$ features. The feature matrix is sparse
but some rows have many nonzero elements so that $\omega=114$.
The URL reputation dataset has a large size: $m=2396130$ examples and $n=3231961$ features. The maximum number of
nonzero elements in a row is $\omega=414$.
We used 16 processors on a computer with Intel Xeon processors at 2.6~GHz and 128~GB RAM.

\begin{figure}
\centering

\includegraphics[width=25em]{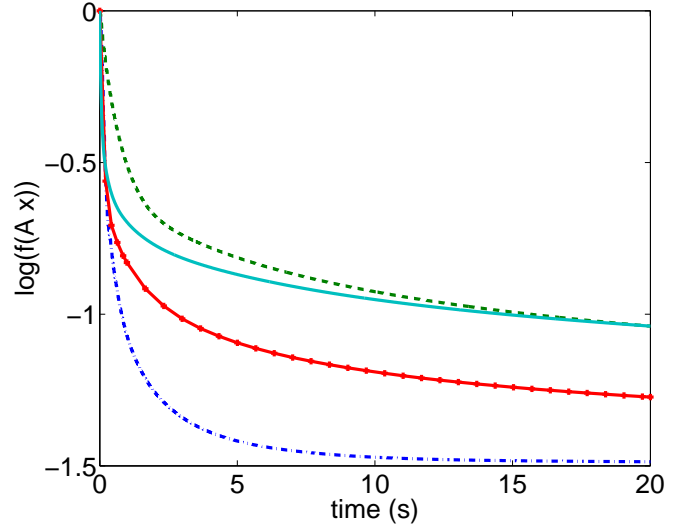}
\caption{Comparison of algorithms for the resolution of the Adaboost problem on the w8a dataset with 16 processors. 
 Dotted line (green): Fully parallel coordinate descent.  Solid line (cyan): Greedy coordinate descent. 
 Solid with crosses (red): Parallel Coordinate Descent with $\tau$-independent sampling ($\tau=16$, $\beta \approx 15.1$). 
Dash-dotted line (blue): Accelerated gradient.
}
\label{fig:expeAdaboost_w8a}
\end{figure}

\begin{figure}
\centering

\includegraphics[width=25em]{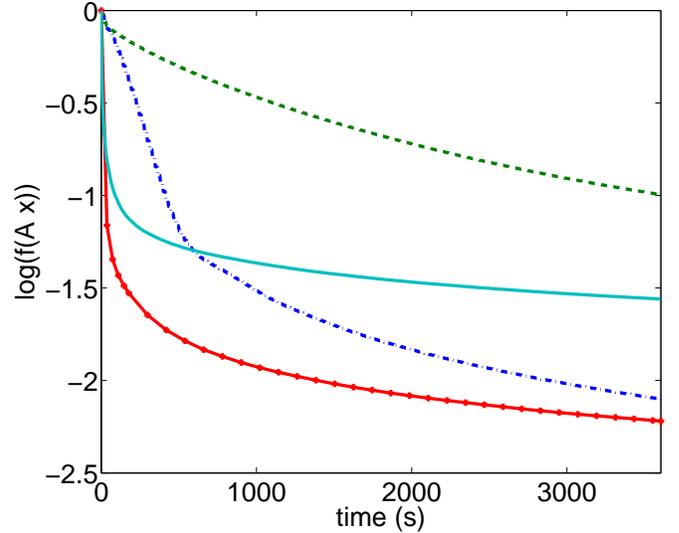}

\caption{Comparison of algorithms for the resolution of the Adaboost problem on the URL reputation dataset with 16 processors
(same colours as in Figure~\ref{fig:expeAdaboost_w8a}). 
}
\label{fig:expeAdaboost_url}
\end{figure}

We give in Figures~\ref{fig:expeAdaboost_w8a} and~\ref{fig:expeAdaboost_url} the value of the objective function at each iteration for:
\begin{itemize}
\item[-] an asynchronous version of Parallel Coordinate Descent with $\tau$-independent sampling ($\tau=16$) (Algorithm~\ref{alg:paralleladaboost}) based on the code of~\cite{RT:PCDM} which is freely available; the $\tau$-independent sampling is a good approximation of the $\tau$-nice sampling for $\tau \ll n$,
\item[-] the fully parallel coordinate descent method~\cite{collins2002logistic},
\item[-] the accelerated version of  the fully parallel coordinate descent method~\cite{mukherjee2013parallel},
\item[-]  the classical Adaboost algorithm (greedy coordinate descent); we performed the search for the largest absolute value of the gradient in parallel.
\end{itemize}

In both cases,
the Parallel Coordinate Descent  with $\tau$-independent sampling is faster than the fully parallel coordinate descent because it benefits from larger steps (the ratio is $\frac{\omega}{\beta}$). It is also faster than the greedy coordinate descent. 
The reason is that one needs to compute the whole gradient at each iteration but only one directional derivative is actually used.   

On the medium scale dataset w8a (Figure~\ref{fig:expeAdaboost_w8a}), the accelerated gradient is the fastest algorithm: it reaches high accuracy in the 20 seconds allocated. 
However, for the large scale dataset URL reputation (Figure~\ref{fig:expeAdaboost_url}), Parallel Coordinate Descent
is the fastest algorithm. It benefits from larger steps ($\beta \approx 3.2$) but also, unlike the other algorithms, 
the computational complexity of one iteration only moderately increases when the size of the problem increases.

\begin{figure}

\centering

\includegraphics[width=25em]{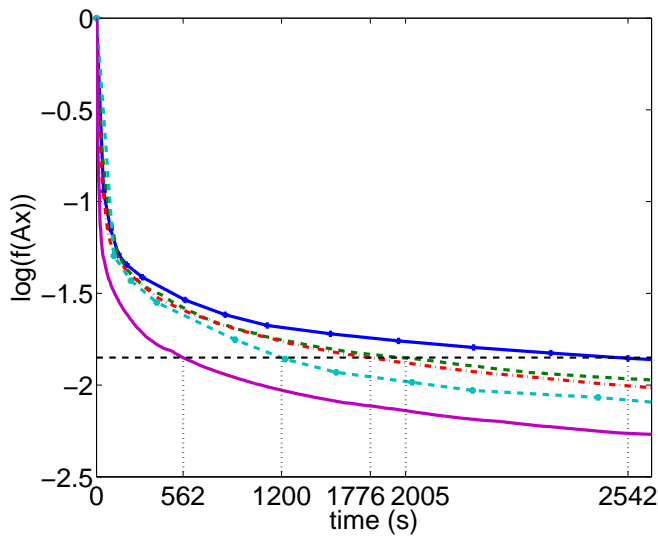}
\caption{
Performance of the smoothed parallel coordinate descent method on
the Adaboost problem for the URL reputation dataset.
Blue solid line with crosses: $\tau=1$; green dashed line: $\tau=2$;
red dash-dotted line: $\tau=4$; cyan dashed line with stars: $\tau=8$;
 purple solid line: $\tau=16$. 
}
\label{fig:tauincreases}
\end{figure}

We can see on Figure~\ref{fig:tauincreases} that increasing the number of
processors indeed accelerates and that the parallelisation speedup factor is
nearly linear: the time needed to reach -1.8 decreases when more processors are used.

\section{Conclusion}

We showed in this paper that the randomised parallel coordinate descent, developed in the 
general framework of the minimisation of composite, partially separable functions, is well suited to solve the Adaboost problem.
We showed that the iteration complexity of the algorithm is of the order $O(1 / \epsilon)$  and gave
a computable formula for the theoretical parallelisation speedup factor. 
The numerical experiments demonstrate the efficiency of  parallel coordinate descent with independent sampling, especially for large scale problems. 
Indeed, each directional derivative computed is actually used to update the optimisation variable  but in the same time  the step length are rather large. The step lengths are controlled by the inverse of the parameter $\beta$ and  
they decrease slower than the increase of the number of processors.
Hence, parallel coordinate descent combines qualities of the greedy coordinate descent and of the fully parallel coordinate descent, so that for large scale problems it outperforms any previously available algorithm.

\bibliographystyle{IEEEtran}
\bibliography{../SmoothingHuge/literature}

\ifthenelse{\boolean{arxiv}}
{

\section*{Appendix: Proof of the iteration complexity}

Let $D^1_K(x)$ be the distance from point $x$ to set $K$ in the $1$-norm:
\[
D^1_K(x)=\min_{z \in K} \norm{x-z}_1\enspace.
\]
We will denote by $P_K^1(x)$ an arbitrary element of $\arg\min_{z \in K} \norm{x-z}_1$.

$\mathbf{E}_t$ will denote the expectation conditional to knowing the previous choices of coordinates $S^0, S^1, \ldots, S^{t-1}$.

\begin{defn}[\cite{telgarsky2012primal}]
$H(A)$ denotes the hard core of $A$: the collection of examples which receive positive
weight under some dual feasible point, a distribution upon which no weak learner is correlated with
the labels. Symbolically,
\[
H(A) := \{  j \in \{1, \ldots, m\} \st \exists \phi \in   \mathrm{Ker}(A^T) \cap \mathbb{R}^m_+ ,\phi_j > 0 \}.
\]
We shall partition $A \in \mathbb{R}^{m \times n}$ by rows into two matrices $A_0 \in \mathbb{R}^{m_0 \times n}$ and $A_{+} \in \mathbb{R}^{m_{+} \times n}$,
where $A_{+}$ has rows corresponding to $H(A)$, and $m_{+} = \abs{H(A)}$.
\end{defn}

\begin{prop}
\label{prop:decrease}
Let us denote $\lambda^{t}$ the $t^{th}$ iterate of Parallel Adaboost (Algorithm~\ref{alg:paralleladaboost}).
For any compact set $K$, let 
\[
\tilde{\gamma}(A,K)=\inf_{\phi \in K \setminus \mathrm{Ker}(A^T)} \frac{\norm{A^T \phi}^*_L}{D^1_{K \cap \mathrm{Ker}(A^T)}(\phi)} \enspace,
\]
If $\nabla f(A \lambda^t) \in K$, then
\begin{align*}
\mathbf{E}_t [F(&\lambda^{t+1})] \leq F(\lambda^t) - \frac{\tau}{2 \beta n} \norm{\nabla F(\lambda^t)}^2_{L^{-1}} \\
& \leq F(\lambda^t) -  \frac{\tau}{2 \beta n} \frac{\tilde{\gamma}(A,K)^2 D^1_{K \cap \mathrm{Ker}(A^T)}(\nabla f(A \lambda^t))^2}{f(A \lambda^t)^2} \enspace,
\end{align*}
where $\beta$ is defined in Theorem~\ref{thm:spcdm}.
\end{prop}
\begin{IEEEproof}
Let $\delta\in \mathbb{R}^n$ be such that $\delta_i=\frac{1}{\beta L_i}\nabla_i F(\lambda^t)$ for all~$i$.
Then we have $\lambda^{t+1}=\lambda^t+ \delta_{[S_t]}$ where $S_t \sim \hat{S}$.

The stopping criterion $\nabla F(\lambda^t) = 0$ grants that for all $t$, $\nabla f(A \lambda^t) \not  \in \mathrm{Ker}(A^T)$:
\begin{align*}
\tilde{\gamma}(A,K)&=\inf_{\phi \in K \setminus \mathrm{Ker}(A^T)} \frac{\norm{A^T \phi}^*_L}{D^1_{K \cap \mathrm{Ker}(A^T)}(\phi)} \\ 
&\leq   \frac{\norm{A^T \nabla f(A \lambda^t)}^*_L}{D^1_{K \cap \mathrm{Ker}(A^T)}(\nabla f(A \lambda^t))}
\end{align*}
\begin{align*}
 \norm{\nabla F(\lambda^t)}_{L^{-1}} &= \norm{\nabla F(\lambda^t)}_{L}^*= \frac{\norm{A^T\nabla f(A \lambda^t)}^*_L}{f(A \lambda^t)}\\
& \geq \frac{\tilde{\gamma}(A,K) D^1_{K \cap \mathrm{Ker}(A^T)}(\nabla f(A \lambda^t))}{f(A \lambda^t)}
\end{align*}

But by Theorem~\ref{thm:spcdm}, the $(t+1)$th iterate of Parallel Adaboost, $\lambda^{t+1}$, satisfies
\[
 \mathbf{E}[F(\lambda^{t+1}+\delta_{[\hat{S}]})] \leq F(\lambda) + \frac{\mathbf{E}[\absS{\hat{S}}]}{n} \Big( \langle \nabla F(x), \delta \rangle + \frac{\beta}{2} \norm{\delta}_L^2 \Big) \enspace,
\]
which implies by definition of $\hat{S}$ and $\delta$ that 
\begin{align*}
\mathbf{E}_t [&F(\lambda^{t+1})] \leq  F(\lambda^t) - \frac{\tau}{2 \beta n} \norm{\nabla F(\lambda^t)}^2_{L^{-1}}\\
& \leq F(\lambda^t) -  \frac{\tau}{2 \beta n} \frac{\tilde{\gamma}(A,K)^2 D^1_{K \cap \mathrm{Ker}(A^T)}(\nabla f(A \lambda^t))^2}{f(A \lambda^t)^2} \enspace.
\end{align*}

\vspace{-4ex}

\end{IEEEproof}

\begin{prop}
\label{prop:boundFinitePart}
Let $A$ and a compact set $K$ such that $\nabla f(K) \cap \mathrm{Ker}(A^T) \not = \emptyset$ be given.
Then $\log\circ f$ is strongly convex over $K$ and taking $\tilde{c}$ to be the modulus of strong convexity,
for any $x \in K \cap \mathrm{Im}(A)$, 
\[
\log(f(x))- \log(\bar{f}_A) \leq \frac{2}{\tilde{c}} \frac{1}{f(x)^2} \mathrm{D}^1_{\nabla f(K) \cap \mathrm{Ker}(A^T)}(\nabla f(x))^2
\] 
\end{prop}

\begin{IEEEproof}
The optimisation problem
\[
\inf_{x\in K} \inf_{\phi \in \mathbb{R}^m \st \norm{\phi}_2=1} \langle \nabla^2 f(x) \phi , \phi\rangle
\]
attains its minimum by compacity of the feasible set and continuity of the objective function. 
\[
\langle \nabla^2 f(x) \phi , \phi\rangle = \sum_{j=1}^m \phi_j^2 q_j(x) - (\sum_{j=1}^m \phi_j q_j(x))^2
\]
where 
\[
q_j(x)=\frac{e^{x_j}}{\sum_{k=1}^m e^{x_k}}>0  \text{ and } \sum_{j=1}^m q_j(x)=1\enspace,
\]
so by Cauchy-Schwarz, $ \sum_{j=1}^m \phi_j^2 q_j(x) \sum_{j=1}^m q_j(x) \geq (\sum_{j=1}^m \phi_j q_j(x))^2$ and there is equality if and only if $\phi_j^2 q_j(x)=q_j(x)$ for all $j$.
Hence, the objective is zero if and only if  $\phi_j^2 q_j(x)=q_j(x)$ for all $j$, which would imply $\phi_j^2=1$ for all $j$ and $\norm{\phi_j}=\sqrt{m}>1$.
We conclude that the optimal value, which is the modulus of strong convexity $\tilde{c}$, is positive.

For the second part of the proposition, we remark that
$\mathrm{Ker(A^T)}$ is a linear space, so $\nabla f(K) \cap \mathrm{Ker(A^T)} \not = \emptyset \Rightarrow \nabla (\log \circ f)(\mathcal{C}) \cap \mathrm{Ker(A^T)} \not = \emptyset$
and we can replace in the proof of Lemma 6.8 in \cite{telgarsky2012primal} $f$ by $\log \circ f$ to get
\begin{multline*}
\log(f(x)) - \log(\bar{f}_A)\\  \leq \frac{1}{2\tilde{c}} \inf_{\psi \in \nabla (\log \circ f)(K) \cap \mathrm{Ker}(A^T)} \norm{\nabla (\log \circ f)(x)-\psi}_1^2
\end{multline*}
We continue by noting that, as $f(x)=\norm{\nabla f(x)}_1$, if $\psi \in \nabla f(K)$, then $\frac{\psi}{\norm{\psi}_1} \in \nabla (\log \circ f)(K)$:
 \begin{align*}
\log&(f(x)) - \log(\bar{f}_A) \\
&\leq \frac{1}{2\tilde{c}} \inf_{\psi \in \nabla f(K) \cap \mathrm{Ker}(A^T)} \norm{ \frac{\nabla f(x)}{ f(x)}-\frac{\psi}{\norm{\psi}_1} }_1^2 \\
&\leq  \frac{1}{2\tilde{c}}  \inf_{\psi \in \nabla f(K) \cap \mathrm{Ker}(A^T)} \Big( \norm{ \frac{\nabla f(x)}{f(x) }-\frac{\psi}{f(x)}}_1 \\ 
& \qquad +\norm{\psi}_1  \abs{\frac{1}{\norm{\psi}_1} - \frac{1}{f(x)}} \Big)^2 \\
&=  \frac{1}{2\tilde{c}}  \inf_{\psi \in \nabla f(K) \cap \mathrm{Ker}(A^T)} \Big( \norm{ \frac{\nabla f(x)}{f(x) }-\frac{\psi}{f(x)}}_1 \\ & \qquad+ \frac{1}{f(x)}
\abs{f(x) - \norm{\psi}_1 } \Big)^2 \\
&\leq  \frac{1}{2\tilde{c}}  \Big( \frac{1}{f(x) }\norm{ \nabla f(x)-P^1_{ \nabla f(K) \cap \mathrm{Ker}(A^T)} (\nabla_f(x)) }_1^2  \\
& \qquad+ \frac{1}{f(x)}
\Big\lvert \norm{ \nabla f(x)}_1 - \norm{P^1_{ \nabla f(K) \cap \mathrm{Ker}(A^T)}(\nabla f(x))}_1 \Big \rvert \Big)^2 \\
&\leq \frac{2}{ \tilde{c} }  \frac{1}{f(x)^2 }\norm{ \nabla f(x)-P^1_{ \nabla f(K) \cap \mathrm{Ker}(A^T)}(\nabla f(x)) }_1^2 
\end{align*}
The last inequality uses the fact that the difference of norms is smaller than the norm of the difference.
The result follows by definition of $P^1_{ \nabla f(K) \cap \mathrm{Ker}(A^T)}(\nabla f(x))$.
\end{IEEEproof}

\begin{IEEEproof}[Proof of Theorem~\ref{thm:convergence}]
We follow the lines of Telgarsky's~\cite{telgarsky2012primal} for the proof. The main differences are the 2-norm instead of Inf-norm in the definition of problem-dependent quantities and the stochastic sampling instead of deterministic sampling.

Theorem~5.9 in \cite{telgarsky2012primal} is still valid in our context, so that $\bar{f}_{A_+}=\bar{f}_A$ and the form of
$f$ gives $f(A \lambda^t)=f(A_0 \lambda^t)+f(A_+ \lambda^t)$.
Thus,
\begin{align*}
F(\lambda^t)-\bar{F}&=\log(f(A_0 \lambda^t)+f(A_+ \lambda^t)) - \log(\bar{f}_A) \\
&=\log (f(A_+ \lambda^t)) + \log\Big(1+\frac{f(A_0\lambda^t)}{f(A_+ \lambda^t)}\Big) - \log(\bar{f}_{A_+}) \\
&\leq \frac{f(A_0\lambda^t)}{f(A_+ \lambda^t)}+ \log (f(A_+ \lambda^t)) - \log(\bar{f}_{A_+})
\end{align*}

For the left term,
\begin{align*}
f(A_0\lambda^t)
=&\norm{\nabla f (A_0 \lambda^t)}_1 
= \norm {\nabla f(A_0 \lambda^t) - \mathrm{P}_{\Phi_{A_0}}^1(\nabla f(A_0 \lambda^t))}_1 \\
= &D^1_{\Phi_{A_0}}(\nabla f(A_0 \lambda^t))
\end{align*}
where we used the fact that $\Phi_{A_0} = 0_{\mathbb{R}^{m_0}}$ (Theorem 5.9 in \cite{telgarsky2012primal}).
Hence
\[
\frac{f(A_0 \lambda^t)}{f(A_+ \lambda^t)} \leq \frac{ D^1_{\Phi_{A_0}}(\nabla f(A_0 \lambda^t))}{f(A_+ \lambda^t)}
\]

For the right term, as in \cite{telgarsky2012primal}, using the fact that the objective values never increase with Parallel Adaboost,
\[
f(A \lambda^0) \geq f(A\lambda^t) \geq f(A_+ \lambda^t) \enspace.
\]
Let $K_+=\{ x \in \mathbb{R}^{m_+} \st (f+I_{Im(A_+)})(x) \leq f(A \lambda^0) \} = \{ x \in \mathbb{R}^{m_+} \st (\log \circ f +I_{Im(A_+)})(x) \leq F(\lambda^0) \}$.
As the level sets of $f$ and $\log \circ f$ are equal, one can reuse the argument on 0-coecivity in \cite{telgarsky2012primal} for
$\log \circ f$. Hence there exists an axis-aligned rectangle $\mathcal{C}_+ \subseteq \mathbb{R}^{m_+}$ containing $K_+$ and such that $\nabla f(\mathcal{C}_+) \cap \mathrm{Ker}(A^T_+) \not = \emptyset$. Moreover, by Proposition \ref{prop:boundFinitePart},
\begin{align*}
\log(f(A_+\lambda^t&))- \log(\bar{f}_{A_+}) \\ 
&\leq \frac{2}{\tilde{c}} \frac{1}{f(A_+ \lambda^t)^2} D^1_{\nabla f(\mathcal{C}_+) \cap \mathrm{Ker}(A^T)}(\nabla f(A_+ \lambda^t))^2.
\end{align*}

We can now merge both estimates as in \cite{telgarsky2012primal}, using Lemma G.2 of \cite{telgarsky2012primal}:
\begin{align*}
 \log(f(A\lambda^t))- \log(\bar{f}_{A}) \leq  \log(f(A_+\lambda^t))- \log(\bar{f}_{A_+}) + \frac{f(A_0 \lambda^t)}{f(A_+ \lambda^t)} \\
\leq \frac{1+2\tilde{w}/\tilde{c}}{f(A_+ \lambda^t)} \mathrm{D}^1_{\mathbb{R}_+^{m_0} \times \nabla f(\mathcal{C}_+) \cap \mathrm{Ker}(A^T)}(\nabla f(A \lambda^t))  \enspace .
\end{align*}

Let $K= \mathbb{R}_+^{m_0} \times \nabla f(\mathcal{C}_+)$. Combining this with Proposition~\ref{prop:decrease}, we get

\begin{align*}
\mathbf{E}_t [F(&\lambda^{t+1})-\bar{F}] - (F(\lambda^t) -\bar{F})\leq - \frac{\tau}{2 \beta n} \norm{\nabla F(\lambda^t)}^2_L  \\
&\leq-  \frac{\tau}{2 \beta n} \frac{\tilde{\gamma}(A,K)^2 D^1_{K \cap \mathrm{Ker}(A^T)}(\nabla f(A \lambda^t))^2}{f(A \lambda^t)^2} \\
&\leq -\frac{\tau}{2 \beta n} \frac{\tilde{\gamma}(A,K)^2 D^1_{K \cap \mathrm{Ker}(A^T)}(\nabla f(A \lambda^t))^2}{f(A_+ \lambda^t)^2} \frac{f(A_+ \lambda^t)^2}{f(A \lambda^t)^2} \\
&\leq - \frac{\tau}{2 \beta n} \frac{\tilde{\gamma}(A,K)^2 (F(\lambda^t)-\bar{F})^2}{(1+2\tilde{w}/\tilde{c})^2} \frac{\bar{f}_A^2}{f(A \lambda^0)^2}
\enspace.
\end{align*}

Theorem 1 in~\cite{RT:UCDC} with
$c_{RT}= \frac{2 \beta n}{\tau} \frac{(1+2\tilde{w}/\tilde{c})^2}{\tilde{\gamma}(A,K)^2} \frac{f(A \lambda^0)^2}{\bar{f}_A^2}$,
implies that, given $\rho>0$ and $\epsilon>0$, if 
\[
T \geq  \frac{2 \beta n}{\tau} \frac{(1+2\tilde{w}/\tilde{c})^2}{\tilde{\gamma}(A,K)^2} \frac{f(A \lambda^0)^2}{\bar{f}_A^2}\frac{1}{\epsilon}(1 + \log \frac{1}{\rho})+2 
\]
then 
\[
\mathbf{P}( F(\lambda^T) -\bar{F}\leq \epsilon) \geq 1-\rho \enspace.
\]
But for $\epsilon \leq 1$,
\begin{align*}
1-\rho&\leq \mathbf{P}\big( F(\lambda^T) -\bar{F}\leq \epsilon\big) = \mathbf{P}\Big( \log\big(\frac{f(A \lambda^T)}{ \bar{f}_A} \big) \leq \epsilon \Big) \\
&=\mathbf{P}\big( f(A \lambda^T) \leq  \bar{f}_A \exp( \epsilon) \big) \\
&\leq \mathbf{P}\big( f(A \lambda^T) \leq  \bar{f}_A (1+ 2 \epsilon) \big)\\
& =  \mathbf{P}\big( f(A \lambda^T) -  \bar{f}_A \leq 2 \epsilon \bar{f}_A \big) 
\end{align*}
We take $\epsilon'=2 \bar{f}_A \epsilon\leq 2 \bar{f}_A $ and we get the result.
\end{IEEEproof}

We also have all the tools to state the convergence results in the case of weak learnability (Proposition \ref{prop:weaklearn}) and attainability (Proposition \ref{prop:attain}): the proofs are easy adaptations of~\cite{telgarsky2012primal}.
}
{}
\end{document}